\def\year{2019}\relax
\documentclass[letterpaper]{article} 
\usepackage{aaai19}  
\usepackage{times}  
\usepackage{helvet}  
\usepackage{courier}  
\usepackage{url}  
\usepackage{graphicx}  
\usepackage{amsmath}
\usepackage{mathrsfs}
\usepackage{amssymb}
\usepackage{algorithm}
\usepackage{algorithmic}
\floatname{algorithm}{Algorithm}
\usepackage{subfigure}
\usepackage{color}
\usepackage{booktabs}
\usepackage{float}
\usepackage{comment}
\title{Cycle-SUM: Cycle-consistent Adversarial LSTM Networks for Unsupervised Video Summarization}
\author{Li Yuan\textsuperscript{\rm 1},\, Francis EH Tay\textsuperscript{\rm 1},\, Ping Li\textsuperscript{\rm 2},\, Li Zhou\textsuperscript{\rm 1},\, Jiashi Feng\textsuperscript{\rm 1}\\\\
\textsuperscript{\rm 1}National University of Singapore\\
\textsuperscript{\rm 2}Hangzhou Dianzi University\\
{\{ylustcnus,zhouli2025\}@gmail.com},\,
 {\{mpetayeh,elefjia\}@nus.edu.sg}, \,
 {lpcs@hdu.edu.cn}
}

\begin{document}
\maketitle
\begin{abstract}
In this paper, we present a novel unsupervised video summarization model that requires no manual annotation. The proposed model termed Cycle-SUM adopts a new cycle-consistent adversarial LSTM architecture that can effectively maximize the information preserving and  compactness of the summary video. It consists of a frame selector and a cycle-consistent learning based evaluator. The selector is a bi-direction LSTM network that learns video representations that embed the long-range relationships among video frames. The evaluator defines a learnable information preserving metric between original video and summary video and ``supervises'' the selector to identify the most informative frames to form the summary video.
In particular, the evaluator is composed of two generative adversarial networks (GANs), in which the forward GAN is learned to reconstruct original video from summary video while the backward GAN learns to invert the processing. The consistency between the output of such cycle learning is adopted as the information preserving metric for video summarization. We demonstrate the close relation between mutual information maximization and such cycle learning procedure. Experiments on two video summarization benchmark datasets validate the state-of-the-art performance and superiority of the Cycle-SUM model over previous baselines.
\end{abstract}
\section{Introduction}

\begin{figure}[!tp]
\centering
\includegraphics[width=1\columnwidth]{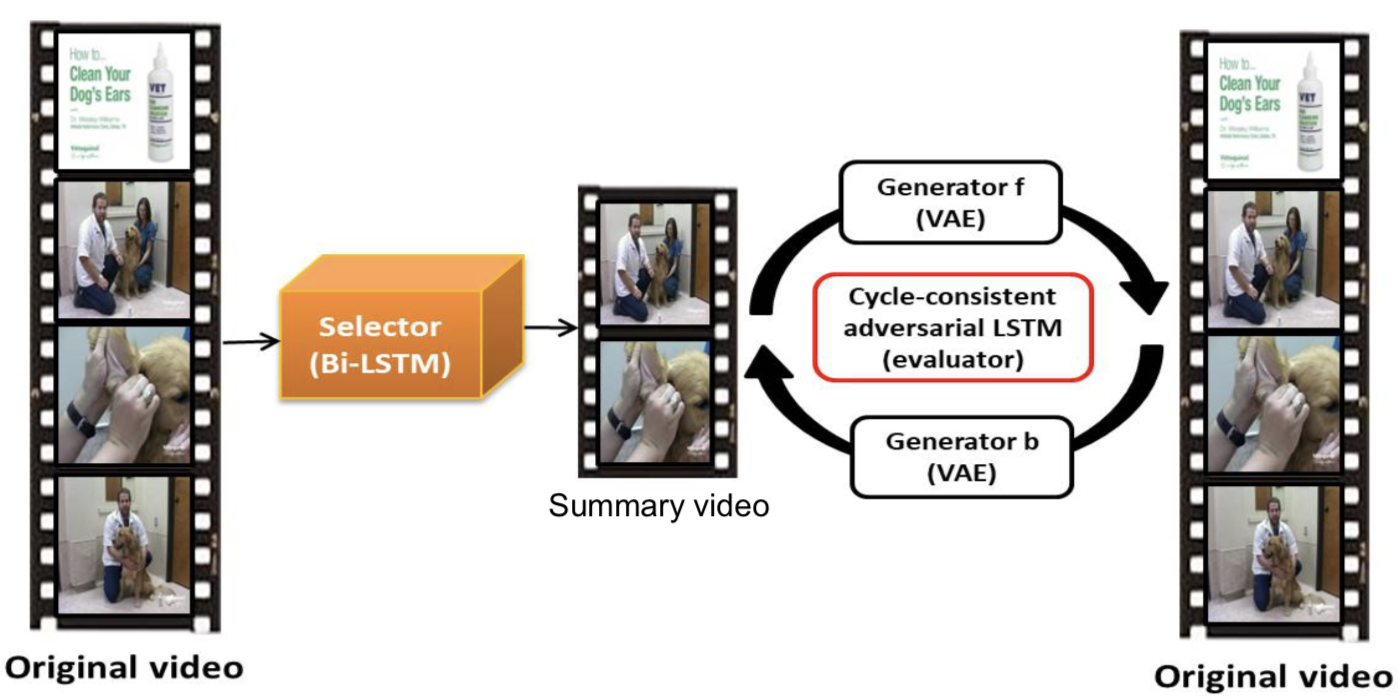}  
\caption{\footnotesize Overview of the Cycle-SUM model. The summary video is selected by the selector from the input original video. To optimize the selector, a cycle-consistent adversarial LSTM evaluator is introduced to evaluate the summary quality through cycle-consistent learning to measure the mutual information between the original and summary video.}
\label{fig:illustration}
\end{figure}


With explosion of video data, video summarization technologies~\cite{ma2002user,pritch2007webcam,lu2013story} become increasingly attractive to help efficiently browse, manage and retrieve video contents. With such techniques, a long video can be shortened to different forms, e.g. key shots~\cite{gygli2014creating}, key frames~\cite{kim2014joint} and key objects~\cite{meng2016keyframes}. Here, we aim at selecting key frames for summarizing a video.


Video summarization is usually formulated as a structure prediction problem~\cite{zhang2016video,mahasseni2017unsupervised}. The model takes as input a sequence of video frames, and outputs a subset of original video frames containing critical information. 
Ideally, the summary video should keep all key information of the original video with minimal redundancy. Summary \emph{completeness} and \emph{compactness} are expected for good video summarization.

Existing approaches can be roughly grouped into supervised and unsupervised ones. Many supervised approaches~\cite{zhang2016video,gygli2015video} utilize human-annotated summary as ground truth to train a model. However, sufficient human-annotated video summarization examples are not always available or expensive to collect. Thus, unsupervised approaches that do not require human intervention become increasingly attractive due to their low cost. For these methods, it is very critical to design a proper summary quality metric. For instance, \cite{mahasseni2017unsupervised} adopt the GAN~\cite{goodfellow2014generative} to measure the similarity between summary and original video and improve the summarization model by optimizing the induced objective, based on a basic idea that a good summary video should be able to faithfully reconstruct the original input video. However, this approach only considers one-direction reconstruction, thus some significant frames may dominate the quality measure, leading to severe information loss in the summary video.

In this paper, we propose a novel cycle-consistent unsupervised model, motivated by maximizing the mutual information between summary video and original video. Our model is developed with a new cycle-consistent adversarial learning objective to pursue optimal information preserving for the summary video, partially inspired by the cycle generative adversarial network~\cite{zhu2017unpaired,yi2017dualgan}. Moreover, to effectively capture the short-range and long-range dependencies among sequential frames~\cite{zhang2016video}, we propose a VAE-based LSTM network as the backbone model for learning video representation.
We name such a cycle-consistent adversarial LSTM network for video summarization as the Cycle-SUM.

Cycle-SUM performs  original and summary video reconstruction in a cycle manner, and leverages consistency between original/summary video and its cycle reconstruction result  to ``supervise'' the video summarization.
Such a cycle-consistent objective guarantees the summary completeness without additional supervision. 
Compared with the one-direction reconstruction (i.e., from summary video to original video)~\cite{zhu2017unpaired,yi2017dualgan}, the bi-direction model performs a reversed reconstruction and a cycle-consistent reconstruction to relieve information loss.

Structurally, the Cycle-SUM model consists of two components: a \emph{selector} to predict an importance score for each frame and select the frames with high importance scores to form the summary video, and a cycle-consistent \emph{evaluator} to  evaluate the quality of selected frames through cycle reconstruction. To achieve effective information preserving, the supervisor employs two VAE-based generators and two discriminators to evaluate the cycle-consistent loss. The forward generator and discriminator are responsible for reconstructing the original video from the summary video, and the backward counterparts perform the backward reconstruction from original to the summary video. Both reconstructions are performed in the learned embedding feature space. The discriminator is trained to distinguish the summary video from original. If the summary video misses some informative frames, the discriminator would tell its difference with the original and thus serves as a good evaluator to encourage the selector to pick important frames.

An illustration of the proposed framework is given in Fig.~\ref{fig:illustration}. The summary video is a subset of all training video frames, selected by the selector based on the predicted frame-wise importance scores. The original video is reconstructed from the summary video, and then back again. Given a distance between original video and summary video in the deep feature space, the Cycle-SUM model tries to optimize the selector such that the distance is minimized over training examples. The closed loop of Cycle-SUM is aimed at 1) assisting the Bi-LSTM selector to select a subset of frames from the original video, and 2) keeping a suitable distance between summary video and original video in the deep features space to improve summary completeness and reduce redundancy.


Our contributions are three-fold.
1) We introduce a new unsupervised video summarization model that does not require any manual annotation on video frame importance yet achieves outstanding performance. 2)
We propose a novel cycle-consistent adversarial learning model. Compared with one-direction reconstruction based models, our model is superior in information preserving and facilitating the learning procedure.
3) We theoretically derive the relation of mutual information maximization, between summary and original video, with the proposed cycle-consistent adversarial learning model. To our best knowledge, this work is the first to transparently reveal how to effectively maximize mutual information by cycle adversarial learning.

\section{Related Work}
Supervised video summarization approaches leverage videos with human annotation on frame importance to train models. For example, Gong et al. formulate video summarization as a supervised subset selection problem and propose a sequential determinantal point processing (seqDPP) based model to sample a representative and diverse subset from training data~\cite{gong2014diverse}. To relieve human annotation burden and reduce cost, unsupervised approaches, which have received increasing attention, generally design different criteria to give importance ranking over frames for selection. For example,   \cite{wang2012event,potapov2014category} propose to select frames according to their content relevance. \cite{mei2015video,cong2012towards} design unsupervised critera by  trying to reconstruct the original video from selected key frames and key shots under the dictionary learning framework. Clustering-based models \cite{de2011vsumm,kuanar2013video} and attention-based models~\cite{ma2002user,ejaz2013efficient} are also developed to select key frames.

Recently, deep learning models are developed for both supervised and unsupervised video summarization, in which LSTM is usually taken as the video representation model. For example,  \cite{zhang2016video} treat video summarization as a sequential prediction problem inspired by speech recognition. They present a bi-direction LSTM architecture to learn the representation of sequential frames in variable length and output a binary vector to indicate which frame to be selected. 
Our proposed Cycle-Sum model also adopts LSTM as backbone for learning long-range dependence between video frames.

\begin{figure*}[!tp]
\centering
\includegraphics[width=2\columnwidth]{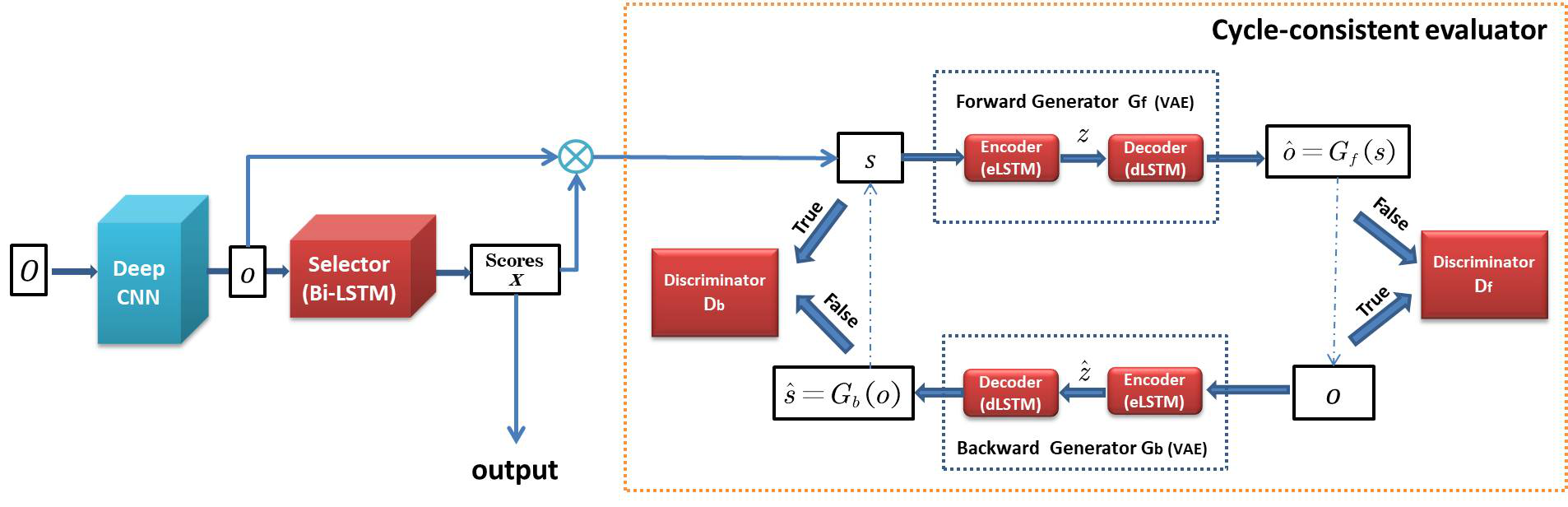}
\caption{\footnotesize Demonstration of Cycle-SUM architecture. Red parts denote the components of our Cycle-SUM model while blue one denotes data-processing. Cycle-SUM has two parts: the selector for selecting frames and the cycle-consistent evaluator to "supervise" the selection. The feature of the frame in original video $o$ is extracted from video $O$ by a deep CNN. The selector takes $o$ as input and outputs the importance scores $x$. During training, the generator $G_{f}$ takes $s$ as input and reconstructs a sequence of features, $G_{f}\left ( s \right )$. The discriminator $D_{f}$ is trained to distinguish $\widehat{o}$ and $o$. The generator $G_{b}$ takes $o$ as input and outputs $G_{b}\left ( o \right )$; the discriminator $D_{b}$ also tries to distinguish between $s$ and $\widehat{s}$. To achieve cycle consistency, the forward cycle $s\rightarrow G_{f}(s)\rightarrow G_{b}(G_{f}(s))\approx s$ and the backward cycle $o\rightarrow G_{b}(o)\rightarrow G_{f}(G_{b}(o))\approx o$ are implemented to encourage the information to be consistent between $o$ and $s$.}
\label{fig:arch}
\end{figure*}
\section{Method}
The proposed Cycle-SUM model formulates video summarization as a sequence-to-sequence learning problem, taking as input a sequence of video frames and outputting a sequence of frame-wise importance scores. The frames with high importance scores are selected to form a summary video. Throughout the paper, we use $O$ and $S$ to denote the original input video and summary video respectively, and $o$ and $s$ to denote the frame-level features of $O$ and $S$ respectively.
To train Cycle-SUM in an unsupervised manner, we develop the cycle-consistent learning method for maximizing mutual information between $o$ and $s$.


\subsection{Mutual Information Maximization via Cycle Learning}
Video summarization is essentially aimed at extracting video frames that contain critical information of the original video. In this subsection, we explain how to derive our cycle-consistent learning objective through the desired objective of maximizing the mutual information between the summary video $s$ and the original video $o$.

Formally, the mutual information $I(o,s)$ is defined as
\begin{equation*}
\begin{aligned}
I(o,s) & \triangleq  \sum_{o}p(o) D_{KL}\left(p(s|o)||p(s)\right),
\end{aligned}
\end{equation*}
where $D_{KL}$ is the KL-divergence between two distributions. Then the objective of video summarization is to extract the summary video $s$ from $o$ to maximize their mutual information. The video summarization model should try to produce $s$ such that its  conditional distribution $p(s|o)$ gives the maximal mutual information with $p(s)$.  However, though it is easy to obtain empirical distribution estimation of original video $o$, it is difficult to obtain ground truth distribution $p(s)$ of corresponding $s$ in an unsupervised learning scenario. This makes one major challenge to unsupervised video summarization.

We propose a cycle-consistent learning objective to relieve such learning difficulty. We notice that
\begin{equation}
\label{eqn:mutual_info}
\begin{aligned}
I(o,s) 
&= \frac{1}{2} \left[\sum_{o}p(o) D_{KL}\left(p(s|o)||p(s)\right) \right. \\
& \qquad \quad \left. + \sum_{s}p(s) D_{KL}\left(p(o|s)||p(o)\right) \right].
\end{aligned}
\end{equation}
The above mutual information computation ``anchors'' at $p(o)$ that can be faithfully estimated and thus eases the procedure of learning distribution of $s$ even in an unsupervised learning setting.

To effectively model and optimize the above learning objective, we adopt the Fenchel conjugate to derive its  bound that is easier to optimize. The Fenchel conjugate of a function $f$ is defined as
$f^*(t) \triangleq \sup_{ u \in \mathrm{dom}_f} \{ut - f(u) \}$,
or equivalently
$f(u) = \sup_{ t \in \mathrm{dom}_{f^*}} \{ut - f^*(t) \}$.

Thus, defining $f(u)=\log u$,  we have the following upper bound for the KL-divergence between distributions $p$ and $q$:
\begin{equation*}
\begin{aligned}
& D_{KL}(p||q) =  -\sum_{x} q(x)\log \frac{p(x)}{q(x)} \\
&= - \sum_{x} q(x) \sup_{t} \left(t \frac{p(x)}{q(x)} - f^*(t)\right)   \\
&= - \sum_{x} q(x) \sup_{t} \left(t \frac{p(x)}{q(x)} +1 + \log (-t) \right)   \\
&\leq -  \sup_{T \in \mathcal{T} } \left( \sum_x p(x)\log T(x) + \sum_x q(x)\log (1-T(x))  \right),
\end{aligned}
\end{equation*}
where $t=T(x)-1$ and $\mathcal{T}$ is an arbitrary class of functions $T: \mathcal{X} \rightarrow \mathbb{R}$. The above inequality is due to the Jensen's inequality and functions $\mathcal{T}$ is only a subset of all possible functions.  Therefore, we have
\begin{equation*}
\begin{aligned}
& D_{KL}(p(s|o)||p(s) ) \\
& \leq  -   \sup_{T \in \mathcal{T}} \left( \sum p(s|o)\log T(s) + \sum p(s)\log(1-T(s)) \right) \\
& \approx - \sup_{T \in \mathcal{T}} \frac{1}{|\mathcal{S}|} \left(\sum_{s \sim p(s|o) } \log T(s) + \sum_{s \sim p(s)} \log(1-T(s))  \right).
\end{aligned}
\end{equation*}
  Here $\mathcal{S}$ is the set of produced summary videos. We can use a generative model to estimate  $p(s|o)$. To this end, we follow the generative-adversarial approach \cite{goodfellow2014generative} and use two neural networks, $G_f$ and $D_f$, to implement sampling and data transformation.  Here $G_f$ is the forward generative model, taking the condition $o$ as input and outputting a sample of summary sample $s$. $D_f$ is the forward discriminative model.  We  learn the generative model $G_f$ by finding a saddle-point of the above objective function, where we minimize w.r.t.\ $G_f$ and maximize w.r.t.\ $D_f$:
\begin{equation}
\label{eqn:gan_minmax_os}
\begin{aligned}
&\min_{G_f}\max_{D_f}  \mathcal{L}(G_f, D_f) \\
&= \frac{1}{|\mathcal{S}|} \left(  \sum_{s\sim G_f(o) } \log  D_f(s) + \sum_{s \sim p(s)} \log (1-D_f(s)) \right).
\end{aligned}
\end{equation}
The above objective is similar to the one of GANs, but the generative model is a conditioned one.  

Similarly, we can obtain the learning objective to optimize the KL-divergence $D_{KL}(p(o|s)||p(o))$ by solving
\begin{equation}
\label{eqn:gan_minmax_so}
\begin{aligned}
&\min_{G_b}\max_{D_b} \mathcal{L}(G_b, D_b) \\
&= \frac{1}{|\mathcal{O}|} \left(  \sum_{o\sim G_{b}(s) } \log  D_{b}(o) + \sum_{o \sim p(o)} \log (1-D_{b}(o)) \right).
\end{aligned}
\end{equation}
Substituting Eqn.~\eqref{eqn:gan_minmax_os} and Eqn.~\eqref{eqn:gan_minmax_so} into Eqn.~\eqref{eqn:mutual_info} gives the following cycle learning objective to maximize the mutual information between the original and summary video:
\begin{equation}
\label{eqn:gan_minmax}
\begin{aligned}
& \min_{G_f, G_b}\max_{D_f, D_b} \mathcal{L}(G_f, G_b, D_f, D_b) \\
& = \sum_{s\sim G_f(o) } \log  D_f(s) + \sum_{s \sim p(s)} \log (1-D_f(s)) \\
& \qquad + \sum_{o\sim G_{b}(s) } \log  D_{b}(o) + \sum_{o \sim p(o)} \log (1-D_{b}(o)),
\end{aligned}
\end{equation}
where we omit the constant number of original frames. To relieve the difficulties brought by the unknown distribution $p(s)$, we use the following cycle-consistent constraint to further regularize the generative model and the cycle learning processing:
\begin{equation*}
G_b(G_f(o)) \approx o, \text{ and } G_f(G_b(s)) \approx s.
\end{equation*}
We name cycle learning with the above consistent constraint as the cycle-consistent learning.


\subsection{Architecture}

Based on the above derivations, we design the cycle-consistent adversarial model for video summarization (Cycle-SUM). The architecture of our Cycle-SUM model is illustrated in Fig.~\ref{fig:arch}. The selector is a Bi-LSTM network, which is trained to predict an importance score for every frame in the input video. The evaluator consists of two pairs of generators and discriminators. In particular, the forward generator $G_{f}$ and the discriminator $D_{f}$ form the forward GAN; the backward generator $G_{b}$ and the discriminator $D_{b}$ form the backward GAN. The two generators are implemented by variational auto-encoder LSTM, which encode the frame feature to the latent variable $z$ and then decode it to corresponding features.  The two discriminators are LSTM networks that learn to distinguish generated frame features and true features. We extensively use the LSTM architecture here for comprehensively modeling the temporal information across video frames. Moreover, by adopting the joint structure of VAE and GAN, the video similarity can be more reliably measured by generating better high-level representations~\cite{larsen2015autoencoding}. The cycle structure (forward GAN and backward GAN) convert from original to summary video and back again, in which information loss is minimized.

Given a video $O$ of $k$ frames, the first step is to extract its deep features $o=\left \{ o_{t}| t=1,\ldots,k \right \}$ via a deep CNN model. Given these features $o$,  the selector predicts a sequence of importance scores $x=\left \{ x_{t}: x_{t}\in \left [ 0,1 \right] \mid  t=1,\ldots,k \right \}$ indicating the importance level of corresponding frames. During training, the frame feature of summary video $s=\left \{ s_{t} = x_{t}o_{t}| t=1,\ldots,k \right \}$. But for testing, Cycle-SUM outputs discretized importance scores $x=\left \{ x_{t}: x_{t}\in \left \{ 0,1 \right\} \right \}$; then frames with importance scores being 1 are selected.

With $s$ and $o$, the supervisor performs  cycle-consistent learning  (see Fig.~\ref{fig:arch}) to evaluate the quality of summary video $s$ w.r.t. both completeness and compactness.  Specifically, within the selector, the forward generator  $G_{f}$ takes the current summary video  $s$ as input and outputs a sequence of reconstructed features for the original video, namely $G_{f}\left ( s \right )=\left \{ \widehat{o}_{t}| t=1,...,k \right \}$. The paired discriminator $D_{f}$ then estimates the distribution divergence  between original video and summary video in the learned feature space.  The backward generator $G_b$ and discriminator $D_b$  have a symmetrical network architecture and training procedure to the forward ones. In particular, the generator $G_{b}$ takes the original video feature $o$ as input, and outputs $G_{b}\left ( o \right )=\left \{ \hat{s}_{t}: t=1,...,k \right \}$ to reconstruct the summary video. The discriminator $D_{b}$ then tries to distinguish between $s$ and $\widehat{s}$. The forward cycle-consistency processing $s\rightarrow G_{f}(s)\rightarrow G_{b}(G_{f}(s))\approx s$ and the backward cycle-consistency $o\rightarrow G_{b}(o)\rightarrow G_{f}(G_{b}(o))\approx o$ are implemented to enhance the information consistency between $o$ and $s$. This cycle-consistent processing guarantees the original video to be reconstructed from the summary video and vice versa, meaning the summary video can tell the same story as the original.

\subsection{Training Loss}
We design the following loss functions to train our Cycle-SUM model. The sparsity loss $\mathcal{L}_{\text{sparsity}}$ is used to control the summary length for the selector; the prior loss $\mathcal{L}_{\text{prior}}$ and the reconstruction loss $\mathcal{L}_{\text{recon}}$ are used to train the two VAE-based generators; adversarial losses $\mathcal{L}_{\text{GAN}}$ are derived from the forward GAN and backward GAN; and $\mathcal{L}_{\text{cycle}}$ is the cycle-consistent loss.

\paragraph{{Sparsity loss} $\mathcal{L}_{\text{sparsity}}$}

This loss is designed to penalize the number of selected frames forming the summary video over the original video. A high sparsity ratio gives a shorter summary video.  Formally, it is defined as
\begin{equation*}
\label{eqn:sparisty_loss}
\mathcal{L}_{\text{sparsity}}=\left \|\frac{1}{k}\sum_{t=1}^{k}x_{t} -\sigma \right \|_{2}
\end{equation*}
where $k$ is the total number of video frames and $\sigma$ is a pre-defined percentage of frames to select in video summarization. The ground truth of $\sigma$ in standard benchmark is 15\%~\cite{gygli2014creating,song2015tvsum}, but we empirically set $\sigma$ as 30\% for the selector in training~\cite{mahasseni2017unsupervised}.

\subsubsection{\textbf{Generative loss $\mathcal{L}_{\text{gen}}$}}
We adopt VAE as the generator for reconstruction, thus $\mathcal{L}_{\text{gen}}$ contains the prior loss $\mathcal{L}_{\text{prior}}$ and the reconstruction loss $\mathcal{L}_{\text{recon}}$. For the forward VAE (forward generator $G_{f}$), the encoder encodes input features $s$ into the latent variable $z$. Assume $ p_{z}(z)$ is the prior distribution of latent variables, and the typical reparameterization trick is to set $p_{z}(z)$ as Gaussian Normal distribution~\cite{kingma2013auto}. Define $q_{\psi}(z|s)$ as the posterior distribution and $p_{\theta}(s|z)$ as conditional generative distribution for
$s$, where $\psi$ is the parameter of the encoder and $\theta$ is that of the decoder. The objective of the forward generator is
\begin{equation*}
\label{eqn:gen_f}
\mathcal{L}_{\text{gen,f}}=D_{KL}\left(q_{\psi }(z|s)\parallel p_{z}(z)\right)-\mathbb{E}\left[\log(p_{\theta}(s|z))\right],
\end{equation*}
where the first  term is KL divergence for the prior loss:
$\mathcal{L}_\text{prior}=D_{KL}(q_{\psi }(z|s)\parallel p_{z}(z))$.

The second  term is an element-wise metric for measuring the similarity between samples, so we use it as the reconstruction loss $\mathcal{L}_{\text{recon}}$.
The typical reconstruction loss for auto encoder networks is  the Euclidean distance between input and reconstructed output: $\left \| x-\hat{x} \right \|_{2}$. According to \cite{larsen2015autoencoding}, element-wise metrics cannot model properties of human visual perception, thus they propose to jointly train the VAE (the generator) and the GAN discriminator, where hidden representation is used in the discriminator to measure sample similarity. Our proposed Cycle-SUM also adopts the same structure to measure the video distance and achieves a feature-wise reconstruction. Specifically, if $x$ and $\hat{x}$ are the input and output of the VAE (the generator), the output of the last hidden layer of the discriminator are $\phi(x)$ and $\phi(\hat{x})$. Then, consider $p(\phi (x)|e)\propto \textup{exp}(-\left \| \phi (x)-\phi (\hat{x}) \right \|)$. The expectation $\mathbb{E}$ can be computed by empirical average. The reconstruction loss can be re-written as
\begin{equation*}
\label{eqn:reconst_loss}
\mathcal{L}_{\text{recon}}=\mathbb{E}[-\log(p_{\theta}(s|e))]\propto \frac{1}{k} \left\| \phi (o)-\phi (\hat{o}) \right \|_{2}.
\end{equation*}
The backward generator $G_{B}$ has the same $\mathcal{L}_\text{prior}$ and $\mathcal{L}_\text{recon}$ as the forward generator. The only difference is that their input and output are reversed.

\subsubsection{\textbf{Adversarial loss} $\mathcal{L}_{\text{GAN}}$}
The learning objective of the evaluator is to maximize the mutual information between the original and summary video. According to Eqn.~\eqref{eqn:gan_minmax}, the adversarial losses for the forward GAN ($G_{f}$ and $D_{f}$) and the backward GAN ($G_{b}$ and $D_{b}$) are
\begin{equation*}
\begin{aligned}
\mathcal{L}_\text{GAN,f}&=\mathbb{E}[\log D_{f}(o)]+\mathbb{E}[1-\log D_{f}(G_{f}(s))],\\
\mathcal{L}_\text{GAN,b}&=\mathbb{E}[\log D_{b}(s)]+\mathbb{E}[1-\log D_{b}(G_{b}(o))].
\end{aligned}
\end{equation*}

To avoid mode collapse and improve stability of optimization, we use the loss suggested in Wasserstein GAN~\cite{arjovsky2017wasserstein}:
\begin{equation*}
\begin{aligned}
\mathcal{L}_{\text{GAN,f}}&=D_{f}(o)-D_{f}(G_{f}(s)), \\
\mathcal{L}_{\text{GAN,b}}&=D_{b}(s)-D_{b}(G_{b}(o)).
\end{aligned}
\end{equation*}

\subsubsection{\textbf{Cycle-Consistent Loss $\mathcal{L}_{\text{cycle}}$}}
Since we expect summary frame features to contain all the information of original frame features, the original video should be fully reconstructed from them. Thus, when we convert from original to summary video and then back again, we should obtain a video similar to the original one. In this way we can safely guarantee the completeness of the summary video. This processing is more advantageous than the one-direction reconstruction in existing image reconstruction works~\cite{zhu2017unpaired,yi2017dualgan}. Based on such an intuition, we introduce the below cycle-consistent loss. The procedure for forward cycle is $s\rightarrow G_{f}(s)\rightarrow G_{b}(G_{f}(s))\approx s$. The $\mathcal{L}_{\text{cycle, f}}$ is correspondingly defined as
$\label{eqn:cycle_loss_f}
\mathcal{L}_{\text{cycle, f}} = \frac{1}{k}\left \| G_{b}(G_{f}(s))-s \right \|$.
For the backward cycle, the procedure is
$o\rightarrow G_{b}(o)\rightarrow G_{f}(G_{b}(o))\approx o$. So the $\mathcal{L}_{\text{cycle, b}}$ is
$\label{eqn:cycle_loss_b}
\mathcal{L}_{\text{cycle, b}} = \frac{1}{k}\left \| G_{f}(G_{b}(o))-o \right \|$.
We adopt ${L}_{1}$ distance for $\mathcal{L}_{cycle}$, since the $L_{2}$ often leads to blurriness~\cite{larsen2015autoencoding,he2016dual}.

\subsubsection{\textbf{Overall Loss $\mathcal{L}$}}
The overall loss function is the overall objective for training the Cycle-SUM model:
\begin{equation*}
\label{eqn:overall_loss}
\begin{split}
\mathcal{L} =
&\mathcal{L}_{\text{sparsity}}+\lambda_{1}(\mathcal{L}_{\text{GAN,f}}+\mathcal{L}_{\text{GAN,b}})\\
&+\lambda_{2}(\mathcal{L}_{\text{gen,f}}+ \mathcal{L}_{\text{gen,b}})+\lambda_{3}(\mathcal{L}_{\text{cycle,f}}+ \mathcal{L}_{\text{cycle,b}}),
\end{split}
\end{equation*}
where $\lambda_{1}$, $\lambda_{2}$ and $\lambda_{3}$ are hyper parameters to balance adversarial processing, generative processing and cycle-consistent processing.


\paragraph{Training Cycle-SUM}
Given the above training losses and final objective function, we adopt the Stochastic Gradient Variational Bayes estimation~\cite{kingma2014stochastic} to update the parameters in training.
The selector and the generators in the Cycle-SUM are jointly trained to maximally confuse the discriminators. To stabilize the training process, we initialize all parameters with Xavier~\cite{glorot2010understanding} and clip all parameters~\cite{arjovsky2017wasserstein}. The clipping parameter $c$ in this training falls in $[-0.5, 0.5]$. The typical value for the generator iteration per discriminator iteration $n$ is $2\sim5$, which means the generator will iterate $n$ times per discriminative iteration. Algorithm~\ref{alg} summarizes all steps for training Cycle-SUM.

\begin{algorithm}[h]
\caption{ Training Cycle-SUM model}
\label{alg}
\begin{algorithmic}[1]
\REQUIRE
Frame features of the training video: $o$
\ENSURE
Learned parameters for the selector $\Theta_{S}$, the two generators and discriminators: $\Theta_{G_{f}}$, $\Theta_{G_{b}}$, $\Theta_{D_{f}}$, $\Theta_{D_{b}}$
\STATE
Initialize all parameters by using Xavier approach
\STATE \textbf{repeat}
\STATE \quad \textbf{for} i = 1,...,$n$ do
\STATE \qquad $o \rightarrow$ original frame-level features from CNN
\STATE \qquad $s \rightarrow$ selector($o$) \% selected frame-level features
\STATE \qquad $\hat{o} \rightarrow$ $G_{f}(s)$ \% {reconstruction by generator A}
\STATE \qquad $\hat{s} \rightarrow$ $G_{b}(o)$ \% reconstruction by generator B
\STATE \qquad $s_{\text{cycle}} \rightarrow$ $G_{b}(\hat{o})$ \% forward cycle
\STATE \qquad $o_{\text{cycle}} \rightarrow$ $G_{f}(\hat{s})$ \% backward cycle
\STATE \qquad \% Updates using RMSProp
\STATE \qquad $-\triangledown \mathcal{L}_{\text{overall}}  \Rightarrow \left \{ \Theta_{S}, \Theta_{G_{f}}, \Theta_{G_{b}} \right \} $
\STATE \qquad $clip(\left \{ \Theta_{S}, \Theta_{G_{f}}, \Theta_{G_{b}} \right \}, -c, c)$
\STATE \quad \textbf{end for}
\STATE \quad $+\triangledown \mathcal{L}_{dis,f}\Rightarrow  \left \{ \Theta_{D_{f}} \right \}$ \%maximization update
\STATE \quad $clip(\Theta_{D_{f}}, -c, c)$
\STATE \quad $+\triangledown \mathcal{L}_{dis,b}\Rightarrow  \left \{ \Theta_{D_{b}} \right \}$ \%maximization update
\STATE \quad $clip(\Theta_{D_{b}}, -c, c)$

\STATE
\textbf{until} convergence
\end{algorithmic}
\end{algorithm}

\section{Experiment}
\subsection{Experiment Setup}
\subsubsection{Datasets and Protocol}
We evaluate Cycle-SUM on two  benchmark datasets: SumMe~\cite{gygli2014creating} and TVSum~\cite{song2015tvsum}. The SumMe contains 25 videos  ranging from 1 to 7 minutes, with frame-level binary importance scores. The TVSum contains 50  videos downloaded from YouTube, with shot-level importance scores for each video taking constant from 1 to 5.

Following the convention~\cite{gygli2015video,zhang2016video,mahasseni2017unsupervised}, we adopt the F-measure as the performance metric. Given ground truth and produced summary video, we calculate the harmonic mean F-Scores according to precision and recall for evaluation.

For the TVSum dataset, shot-level ground truths are provided while the outputs of Cycle-SUM in testing are frame-level scores. Thus we follow the method in ~\cite{zhang2016video} to convert frame-level evaluation to shot-level evaluation. 


\subsubsection{{Implementation Details}}
For fairness, the frame features used for training our model are the same with \cite{zhang2016video,mahasseni2017unsupervised}. We extract 1024-d frame features from the output of pool5 layer of the GoogLeNet network~\cite{szegedy2015going} which is pre-trained on ImageNet~\cite{deng2009imagenet}.

Each of the two generators in our Cycle-SUM is a VAE-based LSTM network consisting of an encoder and a decoder, which has two-layers with $300$ hidden units per layer. The decoder LSTM which reconstructs the sequence reversely is easier to train \cite{srivastava2015unsupervised}, so the decoders in Cycle-SUM also reconstruct the frame features in a reverse order. The discriminators are LSTM networks followed by a fully-connected network to produce probability (true or false) for the input.
Following the architecture of WGAN~\cite{arjovsky2017wasserstein}, we remove the Sigmoid function in the last layer of the discriminator to make the model easier to train. The selector is a Bi-LSTM network consisting of three layers, each with $300$ hidden units.

The two VAE generators are initialized by pre-training on frame features of the original video. Similar to \cite{mahasseni2017unsupervised}, such an initialization strategy can also accelerate training and improve overall accuracy.

All experiments are conducted for five times on five random splits and we report the average performance.

\subsection{Quantitative Results}

\begin{table}
\setlength{\belowcaptionskip}{1pt}
\begin{center}
\fontsize{8pt}{12pt}\selectfont
\caption{ Comparison on F-scores of Cycle-SUM with other \textbf{unsupervised} learning approaches on SumMe and TVSum.}
\begin{tabular}{l|c|c}
 \toprule
 & SumMe &TVSum \\
 \midrule
 De Avila, et al.~\cite{de2011vsumm}   & 33.7    & -  \\
 Li, et al. ~\cite{li2010multi}&   26.6  & - \\
 Khosla, et al. ~\cite{khosla2013large} &- & 50\\
 Song, et al. ~\cite{song2015tvsum}&   26.6  & 50\\
 SUM-GAN~\cite{mahasseni2017unsupervised}& 39.1  & 51.7  \\
 {Cycle-SUM}& \textbf{41.9}  & \textbf{57.6} \\
 \bottomrule
\end{tabular}
\label{tab:unsupervised}
\end{center}
\end{table}

We compare our Cycle-SUM model with several unsupervised state-of-the-arts in
Tab.~\ref{tab:unsupervised}. One can see that the Cycle-SUM model outperforms all of them by a margin up to 2.8\%. In particular,  Cycle-SUM  outperforms SUM-GAN~\cite{mahasseni2017unsupervised} across the two datasets, clearly demonstrating effectiveness of our proposed cycle-consistent loss.  On TVSum, the performance improvement is over 5.9\%. These results well prove the superior performance of Cycle-SUM for video summarization.

\begin{figure*}[!tp]
\centering
\includegraphics[width=1.8\columnwidth]{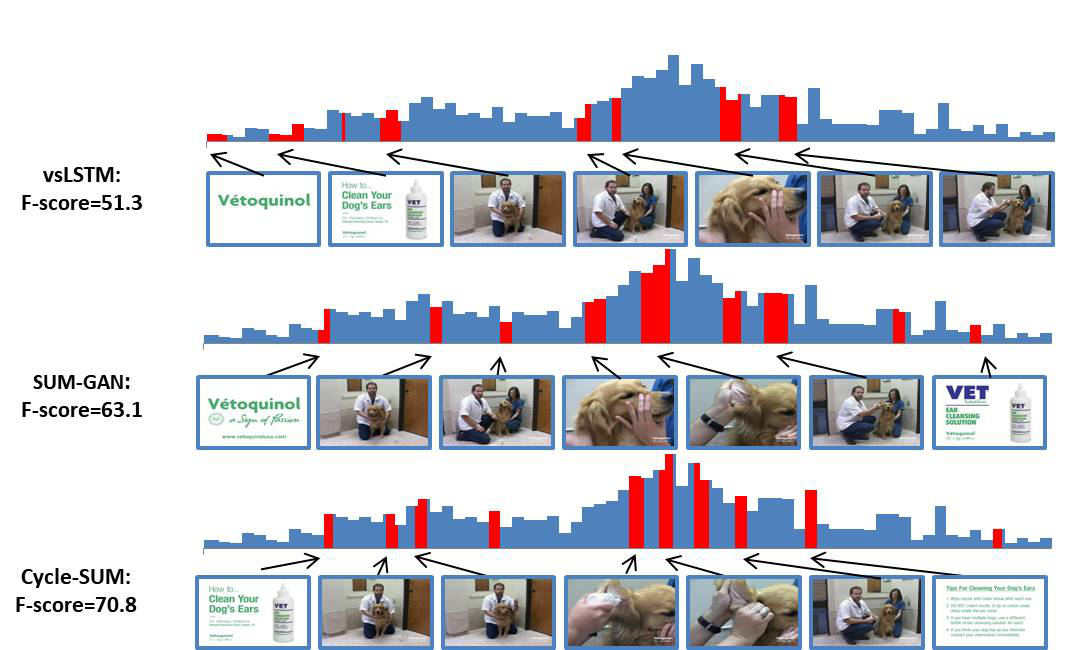}
\caption{\footnotesize Comparison of selected frames w.r.t. importance score by Cycle-SUM and other state-of-arts (vsLSTM and SUM-GAN). Dark blue bars show ground-truth frame-level annotation; Red bars are selected subset shots of all frames. The example video (\# 15) is from TVSum. }
\label{fig:compare_sota}
\end{figure*}

\subsection{Ablation Analysis of Cycle-SUM}

\begin{table}
\setlength{\belowcaptionskip}{3pt}
\begin{center}
\fontsize{8pt}{12pt}\selectfont
\caption{ Performance comparison (F-scores) of the vanilla Cycle-SUM model and its ablation variants on SumMe and TVSum.}
\begin{tabular}{l|c|c}
 \toprule
 & SumMe &TVSum \\
 \midrule
 Cycle-SUM-C &34.8 & 49.5 \\
 Cycle-SUM-1G &38.2 & 51.4\\
 Cycle-SUM-2G &39.7 & 53.6\\
 Cycle-SUM-Gf&40.3 & 55.2\\
 Cycle-SUM-Gb&39.9 & 55.0\\
Cycle-SUM & \textbf{41.9}  & \textbf{57.6}\\
 \bottomrule
\end{tabular}
\label{tab:ablation}
\end{center}
\end{table}

We further conduct ablation analysis to study the effects of different components of the Cycle-SUM model, including the generative adversarial learning and cycle-consistent learning. In particular, we consider following ablation variants of Cycle-SUM.
\begin{itemize}

\item \textbf{Cycle-SUM-C}. This variant is proposed to verify the effects of adversarial learning. It drops the adversarial loss $\mathcal{L}_{\text{GAN,f}}$ and $\mathcal{L}_{\text{GAN},b}$ and keeps all other losses, especially cycle-consistent loss $\mathcal{L}_{\text{cycle}}$ in the overall loss.

\item \textbf{Cycle-SUM-2G}. The cycle-consistent loss $\mathcal{L}_{\text{cycle,f}}$ and $\mathcal{L}_{\text{cycle,b}}$ are removed in overall loss. The forward GAN and backward GAN are still kept. We compare the results of Cycle-SUM-2G with Cycle-SUM to analyze the functions of cycle-consistent reconstruction.

\item \textbf{Cycle-SUM-1G}. The cycle-consistent loss $\mathcal{L}_{\text{cycle,f}}$ and $\mathcal{L}_{\text{cycle,b}}$ are not used when training this variant. Meanwhile, we remove the generator $G_{b}$ and discriminator $D_{b}$, so there is no backward reconstruction: $o\rightarrow G_{b}(o)\rightarrow \hat{s}$. This model is similar to SUM-GAN~\cite{mahasseni2017unsupervised}. It only has forward GAN, and Cycle-SUM-2G contains the two GANs during training. We are also interested in comparing {Cycle-SUM-2G} and {Cycle-SUM-1G}.

\item \textbf{Cycle-SUM-Gf}.
The backward cycle-consistent loss $\mathcal{L}_{\text{cycle,b}}$ is not included in overall objective when training, while forward cycle-consistent loss is still kept. The forward and backward adversarial learning are still kept.

\item \textbf{Cycle-SUM-Gb}.
The forward cycle-consistent loss $\mathcal{L}_{\text{cycle,f}}$ is not included in overall objective when training, and backward one is kept. The forward and backward adversarial learning are still kept.

\end{itemize}




Comparing F-scores of Cycle-SUM and Cycle-SUM-C in Tab.~\ref{tab:ablation}, we can see the adversarial learning improves the performance significantly, proving the positive effects of deploying GAN in unsupervised video summarization. Compared with one GAN variant Cycle-SUM-1G, the results of Cycle-SUM-2G have 2\% gain on average. Both comparisons verify adversarial learning helps improve video summarization.

By comparing Cycle-SUM-2G and Cycle-SUM, we can see  averagely F-scores rise by 2.2\% on SumMe and 4.0\% on TVSum. This proves that cycle-consistent reconstructions can ensure the summary video contain full information of the original video. 

The results of Cycle-SUM-Gf are slightly better than Cycle-SUM-Gb. However, both variants bring performance gain over Cycle-SUM-2G, which also proves  the forward and backward cycle-consistent processing can promote the ability to select a fully summary video from the original.

To sum up, the adversarial learning ensures the summary  and original video to keep a suitable distance in the deep feature space, and the cycle-consistent learning ensures selected  frames to retain full information of the original video.

\subsection{Qualitative Results}

Fig.~\ref{fig:compare_sota} shows summarization examples from a sample video in TVSum. We compare the selected frames of Cycle-SUM with other two recent state-of-the-arts, vsLSTM~\cite{zhang2016video} and SUM-GAN~\cite{mahasseni2017unsupervised} by using a successful example for all three models. 
As shown in Fig.~\ref{fig:compare_sota}, Cycle-SUM selects shorter but more key shots than the other two models.
Compared with  results of vsLSTM and SUM-GAN, some topic-specific and informative details, e.g. frames showing the doctor pushing medicinal liquid into dog's ear, are correctly selected by Cycle-SUM.


\section{Conclusion}
In this paper, we theoretically reveal how to effectively maximize mutual information by cycle-consistent adversarial learning. Based on the theoretical analysis, we propose a new Cycle-SUM model for frame-level video summarization. Experimental results show that the cycle-consistent mechanism can significantly improve video summarization, and our Cycle-SUM can produce more precise summary video than strong baselines, which well validates effectiveness of our method.

\section{ Acknowledgments}
This work was spported in part to Jiashi Feng by NUS IDS R-263-000-C67-646,  ECRA R-263-000-C87-133 and MOE Tier-II R-263-000-D17-112, in part to Ping Li by NSFC under Grant 61872122, 61502131, and in part by the Zhejiang Provincial Natural Science Foundation of China under Grant LY18F020015.

\bibliographystyle{aaai}
\bibliography{aaai19}

\end{document}